\documentclass{article}

\usepackage[nonatbib,final]{nips_2017}

\usepackage[final]{nips_2017}

\usepackage[utf8]{inputenc} 
\usepackage[T1]{fontenc}    
\usepackage{hyperref}       
\usepackage{url}            
\usepackage{booktabs}       
\usepackage{amsfonts}       
\usepackage{nicefrac}       
\usepackage{microtype}      
\usepackage[pdftex]{graphicx}
\usepackage{subfigure}
\usepackage{wrapfig}
\usepackage{amsmath}

\title{Lat-Net: Compressing Lattice Boltzmann Flow Simulations using Deep Neural Networks}

\author{
  Oliver Hennigh \\
  Mexico \\
  \texttt{loliverhennigh101@gmail.com} \\
}

\begin{document}

\maketitle

\begin{abstract}
Computational Fluid Dynamics (CFD) is a hugely important subject with applications in almost every engineering field, however, fluid simulations are extremely computationally and memory demanding. Towards this end, we present Lat-Net, a method for compressing both the computation time and memory usage of Lattice Boltzmann flow simulations using deep neural networks. Lat-Net employs convolutional autoencoders and residual connections in a fully differentiable scheme to compress the state size of a simulation and learn the dynamics on this compressed form. The result is a computationally and memory efficient neural network that can be iterated and queried to reproduce a fluid simulation. We show that once Lat-Net is trained, it can generalize to large grid sizes and complex geometries while maintaining accuracy. We also show that Lat-Net is a general method for compressing other Lattice Boltzmann based simulations such as Electromagnetism.

\end{abstract}

\section{Introduction}

Computational fluid dynamics (CFD) is a branch of fluid mechanics that deals with numerically solving and analyzing fluid flow problems such as those found in aerodynamics, geology, biology, etc. CFD simulations are known for their high computational requirements, memory usage, and run times. Because of this, there is an ever growing body of work using simulation data to create reduced order models or surrogate models that can be evaluated with significantly less resources. Towards this end, we develop a neural network approach that both compresses the computation time and memory usage of fluid simulations.

We investigate fluid simulations that contain complex time dependent turbulence. Simulations of this form are difficult because they require fluid solver to have high resolution and small times steps. Never the less, they frequently occur in nature and are an important area of study. Motivated by need for these simulations and the recent success of neural network based models in related areas \cite{tompson2016accelerating} \cite{guo2016convolutional} \cite{yang2016data}, we choice this setting to test our model.

The most popular approach to modeling fluid flow is with the Navier stokes equation. The solution to this partial differential equation gives the flow velocity field for a given domain. Recently, a new method for simulating fluid flow has emerged named the Lattice Boltzmann Method (LBM). It is derived from the Boltzmann equation and grew out of Lattice Gas Automaton (LGA) in the late 80s \cite{mcnamara1988use}. The main advantages of the LBM are its ability to run on complex geometries, its scalability to parallel architectures (particularly GPUs) and applicability to complex flows that contain phenomena such as heat transfer and chemical reactions. Our method is centered around this method of simulating flow.

Lat-Net works by compressing the state of a simulation while learning the dynamics of the simulation on this compressed form. The model can be broken up into three pieces, an encoder, compression mapping, and decoder. The encoder compresses both the state of the simulation as well as the given boundary conditions. The compression mapping learns the dynamics on the compressed state that correspond to the dynamics in the fluid simulation. The decoder decompresses the compressed state allowing either the whole simulation state or desired pieces to be extracted.

We focus the content of this paper on LBM fluid simulations because this is the most popular use of the LBM, however, this method of simulation is known to be able to solve a large set of partial differential equations \cite{galindolattice}. In fact, LBM can simulate many physical systems of interest such as Electromagnetism, Plasma, Multiphase flow, Schrödinger equation etc. \cite{mendoza2010three} \cite{kim2008wavelet} \cite{zhong2006lattice} \cite{shan1993lattice}. With this in mind, we keep our method general and show evidence our method works equally well on Electromagnetic simulations. However, because the dominate use of LBM is on fluid flow problems we center discussion on this subject.

Our work has the following contributions.
\begin{itemize}
  \item It allows for simulations to be generated with less memory then the original flow solver. There is a crucial need for such methods because memory requirements grow cubic to grid size in 3D simulations. In practice, this quickly results in the need for large GPU clusters \cite{onodera2013large} \cite{xian2011multi}.
  \item Once our model is trained, it can be used to generate significantly larger simulations. This allows the model to learn from a training set of small simulations and then generate simulations as much as 16 times bigger with little effect in accuracy.
  \item Our method is directly applicable to a variety of physics simulations, not just fluid flow. We show this with our electromagnetic example and note that the changes to our model are trivial.
\end{itemize}

\section{Related Work}

Recently, there have been several papers applying neural networks to fluid flow problems. Guo etc. \cite{guo2016convolutional} proposed to use a neural network to learn a mapping from boundary conditions to steady state flow. Most related to our own work, Yang etc. \cite{yang2016data} and Tompson etc. \cite{tompson2016accelerating} use a neural network to solve the Poisson equation in order to accelerate Eulerian fluid simulations. The key difference between this and Lat-Net is its ability to compress the memory usage and the generality of our method to other physics simulations.

There has also been an increasing body of work applying neural networks to other physics modeling problems. For example, neural networks have been readily adopted in many chemistry applications such as predicting molecular properties from descriptors, protein contact prediction and computational material design \cite{goh2017deep}. Very recently, neural networks have been applied to quantum mechanics problems as seen in Mills etc. \cite{mills2017deep} and Giuseppe etc. \cite{carleo2017solving} where neural networks are used to approximate solutions to the Schrödinger equation. In high energy Physics, Paganini etc. \cite{2017arXiv170502355P} uses a generative adversarial networks (GAN)\cite{goodfellow2014generative} to model electromagnetic showers in a longitudinally segmented calorimeter. Many of these applications are relatively recent and indicate a resurgence of interest in applications of neural networks to modeling physics.

Reduced order Modeling is an area of research that focuses on techniques to reduce the dimensionality and computational complexity of mathematical models. A Reduced order model (ROM) is constructed from high-fidelity simulations and can subsequently be used to generate simulations for lower computation. The most popular ROM method for fluid dynamics is Galerikin projection \cite{rowley2004model} \cite{barone2009reduced}. This method uses Proper Orthogonal Decomposition to reduce the dimensionality of flow simulations and then finds the dynamics on this reduced space. There are other methods that build on this such as reduced basis methods and balanced truncation \cite{veroy2005certified} \cite{rowley2005model}. While these approaches are centered around the Navier stokes equation and thus not directly comparable to our own, we note that the compression mapping present in these methods is typically quite simple. Given the recent success neural networks have had in creating well structured encodings (such as Variational Autoencoders \cite{kingma2013auto} \cite{watter2015embed}), we feel our approach is well justified.

\section{Deep Neural Networks for Compressed Lattice Boltzmann}

In this section, we present our model for compressing Lattice Boltzmann simulations.

\subsection{Review: The Lattice Boltzmann Method}

\begin{figure}[!t]
\centering
\subfigure{\includegraphics[scale=0.19]{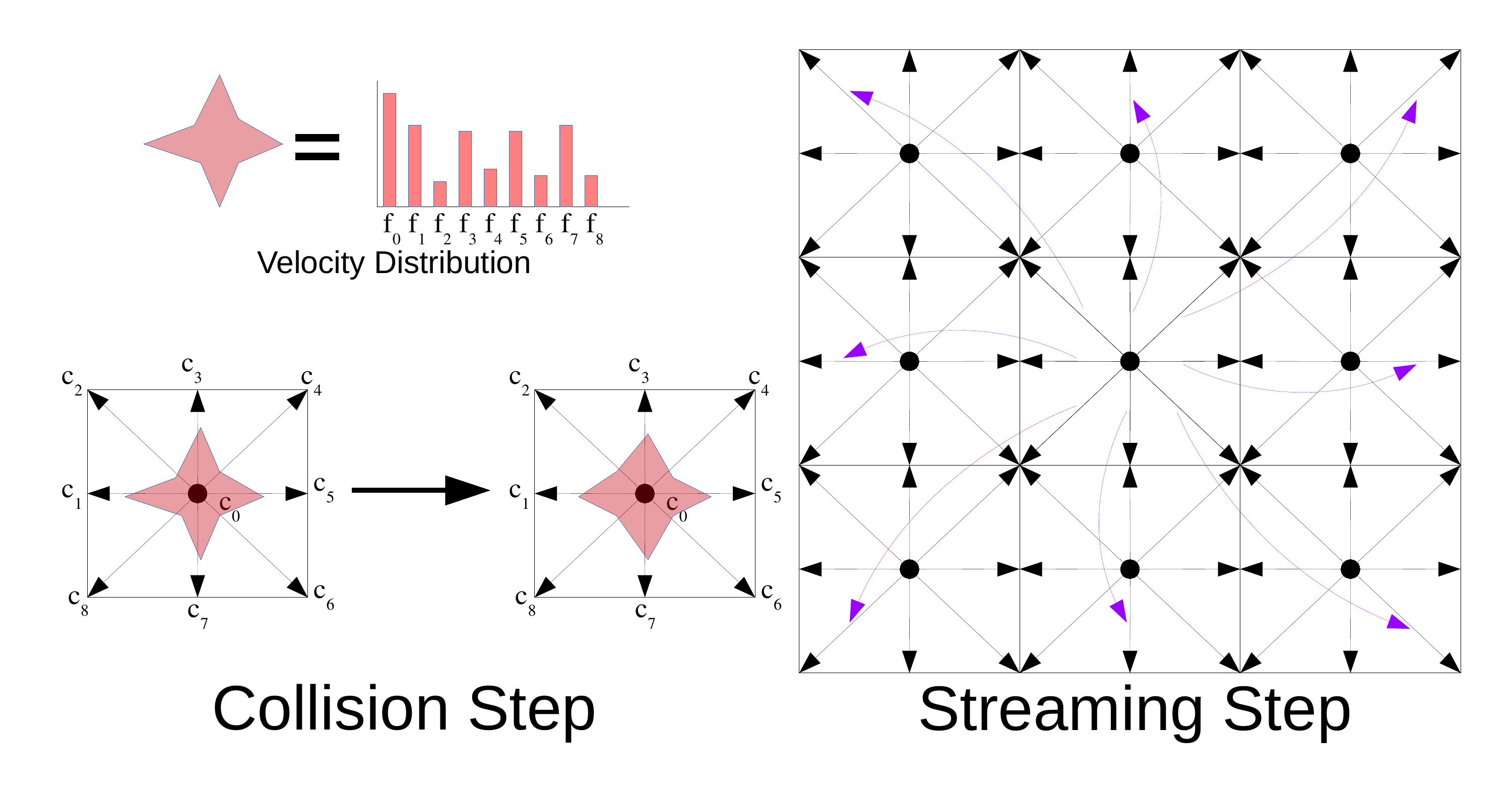}}
\caption{Illustration of the Lattice Boltzmann update steps}
\label{lattice_boltzmann}
\end{figure}

In a Lattice Boltzmann simulation, the domain is discretized into an equal sized Cartesian grid. Each cell of this grid contains a velocity distribution function $f_i$ that describes the velocity of flow at that point. $f_i$ has values ranging over $i$ that correspond to the $\{ \vec{c}_i \}$ directions of flow. In our 2 dimensional simulations, there are 9 such directions (D2Q9 scheme). A figure showing this grid structure is seen in \ref{lattice_boltzmann}. From the distribution function $f_i$, one can calculate the density ($p$) and velocity ($\vec{u}$) of the fluid flow with the following equations.

\begin{equation}
  p = \sum_i f_i  \qquad\text{and}\qquad \vec{u} = \sum_i \vec{c}_i f_i 
\end{equation}

The lattice states are updated with two separate steps, the collision step and the streaming step. The collision step mimics the flow interacting with itself and is updated in the following way,

\begin{equation}
  f^t_i(x, t + \delta_t) = f_i(x,t) + \frac{1}{\tau} (f_i^{eq} - f_i)
\end{equation}

where $\tau$ is the relaxation constant and $f_i^{eq}$ is the flow equilibrium. For our simulations, we use the $f_i^{eq}$ from the Lattice Bhatnagar-Gros-Krook (LBGK) scheme \cite{guo2013lattice}. After the collision step is applied, the flow propagates to adjacent cells following the streaming step. 

\begin{equation}
  f_i(x + c_i, t + \delta_t) = f^t_i(x,t + \delta_t)
\end{equation}

This step will contain bounce back if one of the adjacent cells is a boundary. Figure \ref{lattice_boltzmann} illustrates these steps for the 2 dimensional case.

It is interesting to note the simplicity of this method and its similarity to convolutional neural networks. In fact, if we treat the lattice state $f$ as a tensor of size ($n_x$,$n_y$,9), as we do for the remainder of this paper, the streaming operator can be mimicked with a 3 by 3 convolution and the collision step can be performed with a 1 by 1 convolution (D2Q9). This offers a unique way to interpret our method. In some sense, we are taking a large convolutional neural network and compressing it onto a much smaller and more memory efficient network. With this mental picture in mind, we now describe our approach.

\subsection{Proposed Architecture}

\begin{figure}[!t]
\centering
\subfigure{\includegraphics[scale=0.28]{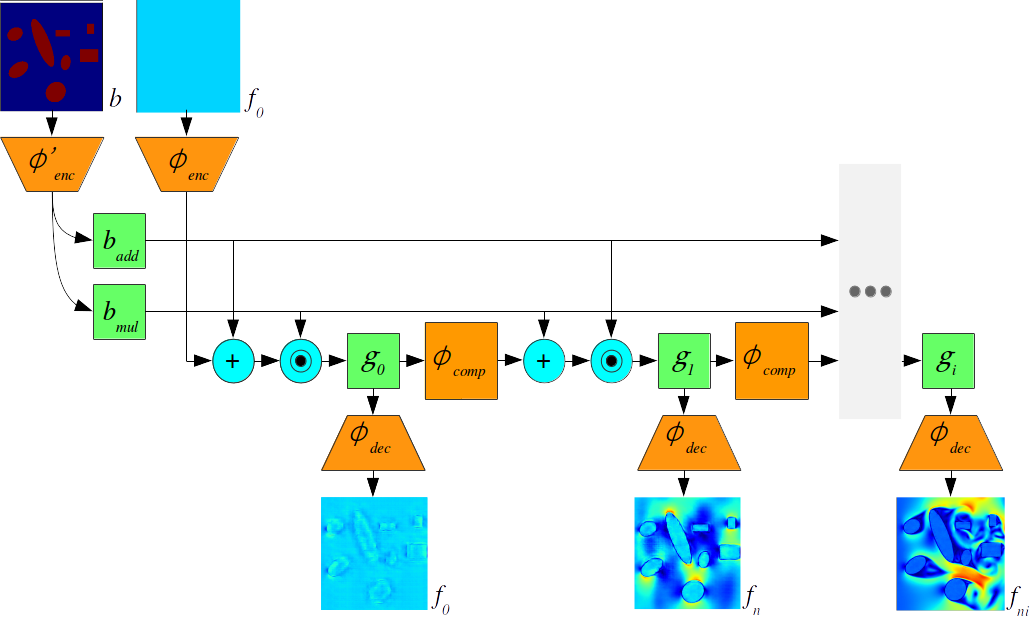}}
\caption{Illustration of the Lat-Net architecture}
\label{fig_1}
\end{figure}

Figure \ref{fig_1} shows a sketch of the model. The figure can be understood by following the arrows starting from the flow state $f_t$ and the boundary $b$. We treat $f_t$ as a tensor with shape ($n_x,n_y,9$) for the 2D case and ($n_x,n_y,n_z,15$) for the 3D case. The boundary is treated as a binary tensor of shape ($n_x,n_y,1$) and ($n_x,n_y,n_z,1$) with the value being 1 if the cell is solid. Bellow we walk through each step of our method.

First, we compress both the state of the fluid simulation $f_t$ and the boundary conditions $b$ using two separate neural networks $\phi_{enc}$ and $\phi'_{enc}$ respectively. The result from $\phi_{enc}$ is a compressed representation of the flow $g_t$ and the result of $\phi'_{enc}$ are two tensors $b_{mul}$ and $b_{add}$ of equal size to $g_t$. These three tensors represent the entirety of the compressed state of the simulation.

In a Lattice Boltzmann solver, the boundary conditions are used at each time-step to add bounce back to the streaming step. In a similar way, our model applies the compressed boundary to the compressed state every time-step. We do this in the following way,
\begin{equation}
  g_t = (g_t \odot b_{mul}) + b_{add}
\end{equation}
This method proved extremely successful at keeping the boundary information firmly planted through the duration of the simulation. This method of applying boundary conditions was inspired by \cite{vondrick2016generating} where they use a similar method to combine foreground and background information in video prediction. After the boundary is applied to $g_t$, we can run the state through another neural network to emulate the dynamics, i.e. $\phi_{comp}:g_{t} \rightarrow g_{t+1}$. Each step of $\phi_{comp}$ is equivalent to $n$ time-steps of the Lattice Boltzmann solver. For example, in the 2 dimensional simulation, each step of $\phi_{comp}$ mimics 120 steps of the Lattice Boltzmann solver. Once $g_t$ is computed, we can extract out the generated state of the simulation with a decoder network $\phi_{dec}$. 

\subsection{Network Implementation Details}

We implement 2 networks trained on 2 dimensional and 3 dimensional lattice simulations. The encoder, compression mapping and decoder pieces of the 2D network are each a series of 3 by 3 residual blocks with the sequences 4x(down res-res)-res, 4x(res), and 3x(transpose conv-res-res)-transpose conv, respectively \cite{he2016deep}. For the 3D network, the sequences are 2x(down res)-res, 3x(res), and (transpose conv-res-transpose conv) where 3 by 3 by 3 convolutions are used. The down residual blocks are created by changing the first convolution to have kernel size 4 by 4, stride 2 and double the filter size. The up sampling is achieved with transpose convolutions of kernel size 4 by 4, stride 2 and half the filter size. For the last residual block on the 3D network, the filter size is halved once again.

As mentioned above, each network is kept entirely convolutional. Fluid flow is inherently spatially correlated so using convlutional layers allows this spatial information to be preserved. Keeping the network convolutional also allows different input sizes to be used. This is how our model is able to train on small simulations and then generate larger simulations.

Residual connections have been used in many deep learning architectures with much success. Adding residual connections allows for much deeper networks to be trained, often resulting in improved results \cite{he2016deep}. When training our model, it is necessary to unroll the compression network over several time-steps. This has the same effect as making the network deeper. For this reason, it seems logical to take advantage of this network architecture. We have seen that removing these residual connections results in much slower convergence and worse accuracy.

\subsection{Training Details}

Lat-Net is trained by unrolling the network and comparing the generated flow with the true. Our loss function is Mean Squared Error (MSE) with Image Gradient Difference Loss (GDL) \cite{mathieu2015deep}. The GDL is multiplied by $\lambda_{GDL}$ and then added to the MES. In all our experiments, $\lambda_{GDL}$ is set to $0.2$. Removing the GDL tended to produce less accurate models. Lat-Net is unrolled 5 time-steps and then trained with the Adam optimizer \cite{kingma2014adam}.

\section{Experiments}

In this section, we describe our experiments testing Lat-Net on a variety of problems. Our experiments are designed to test our models ability to generate large simulations as well as its transferability to new boundary geometries. We also explore computation time and working memory usage. Finally, we briefly show results applying this method to electromagnetic simulations.

\subsection{Dataset Generation}
In order to train and test our model, we generate sets of fluid and electromagnetic simulations. All simulations were generated with the MechSys library \cite{mechsys}.

The train set for the 2D fluid simulations consists of 50 runs of grid size 256 by 256 and 9 directional flows in the lattice Boltzmann solver (D2Q9 scheme)\cite{guo2013lattice}. The simulations use periodic boundary conditions on top and bottom as well as uniform inlet flow and outlet flow of 0.04 from the left and right. 8 Objects are placed randomly with height and width sizes ranging from 140 to 20 cells. The test set for the 2 dimensional simulations consists of 10 runs of size 256 by 256, and 5 runs of size 1024 by 1024 with the same boundary conditions and object densities. We also generate a test set of size 256 by 512 with vehicle cross sections as objects. There are 28 cross sections used ranging from trucks to minivans. For all 2 dimensional simulations, the ratio of network steps to Lattice Boltzmann steps is 1 to 120.

The train set for the 3D fluid simulations consists of 50 runs of grid size 40 by 40 by 160 and 15 directional flows in the lattice Boltzmann solver (D3Q15 scheme)\cite{guo2013lattice}. Similar to the 2D simulations, periodic boundary conditions are used with same inlet and outlet flow. 4 spheres are randomly placed with height and width 24. The reason different object geometries and sizes were not explored was due to the fact that smaller objects or objects with complex geometries tended to have too course a resolution for the lattice Boltzmann solver and larger objects required too large a simulation size. The test set comprises 10 runs of 40 by 40 by 160 and 5 runs of 160 by 160 by 160 simulations with the same object density. The ratio of network steps to Lattice Boltzmann steps is 1 to 60.

The train set for the electromagnetic simulations are grid size 256 by 256 with periodic boundaries. An electromagnetic wave is initialized in the top of the simulations and proceeds to interact with randomly placed objects of different dielectric constants. When the wave hits these objects, the reflection and refraction phenomenon is seen. The test set consists of simulations of size 512 by 512 with the same object density.

\subsection{Generating Simulations}

\begin{figure}[!t]
\centering
\subfigure{\includegraphics[scale=0.302]{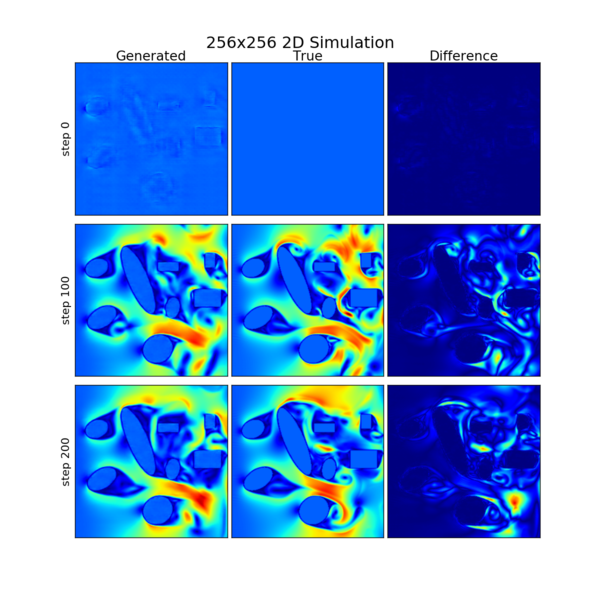}}
\subfigure{\includegraphics[scale=0.302]{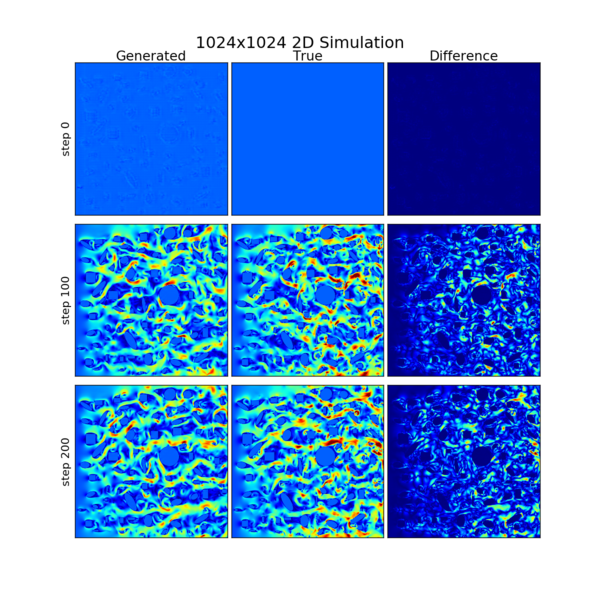}}
\subfigure{\includegraphics[scale=0.302]{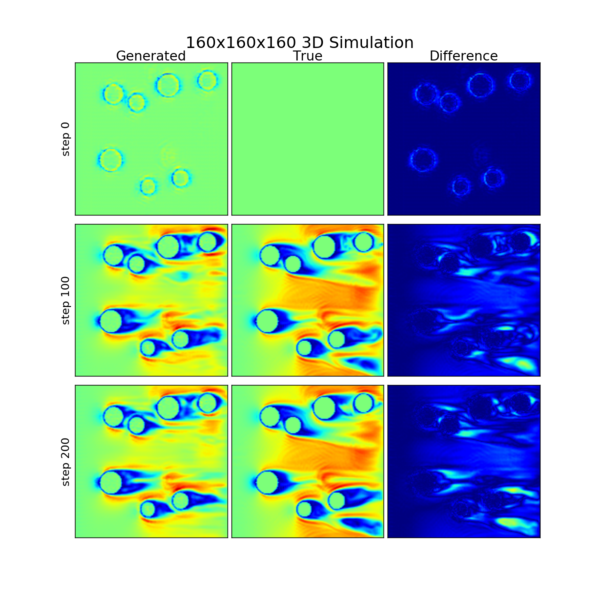}}
\caption{A visual comparison of flows generated by Lat-Net and the Lattice Boltzmann method. Each figure shows the Generated, True, and Difference of the flow for various time-steps.}
\label{2d_image_plot}
\end{figure}

\begin{figure}[!t]
\centering
\subfigure{\includegraphics[scale=0.280]{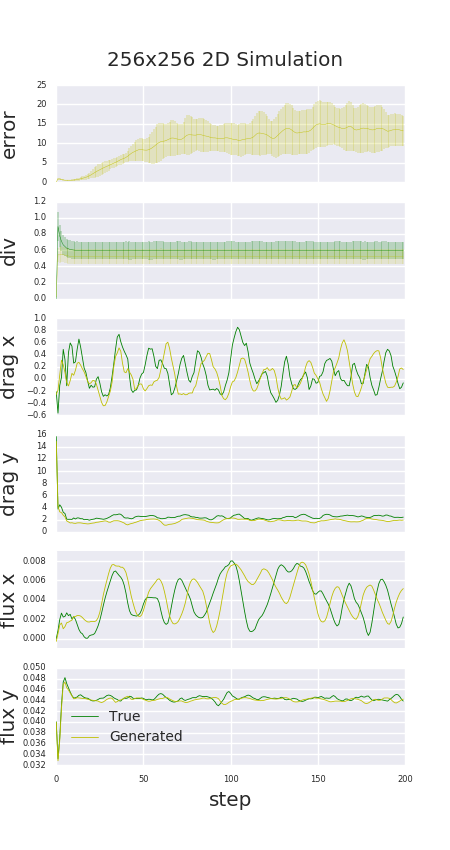}}
\subfigure{\includegraphics[scale=0.280]{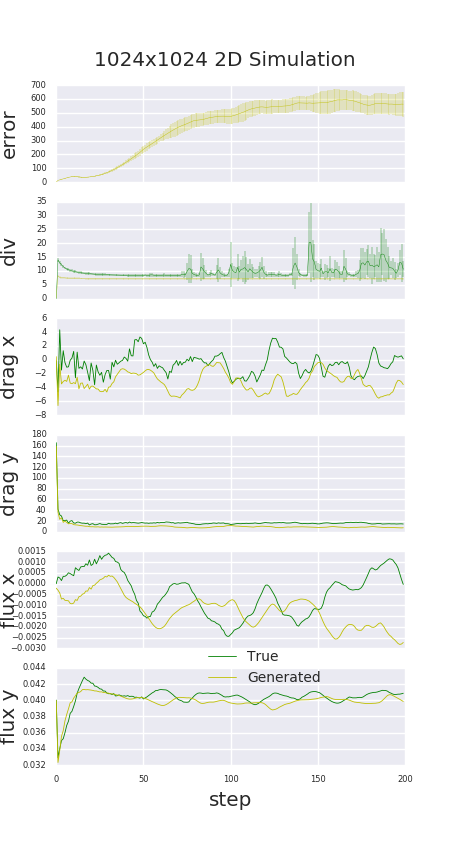}}
\subfigure{\includegraphics[scale=0.280]{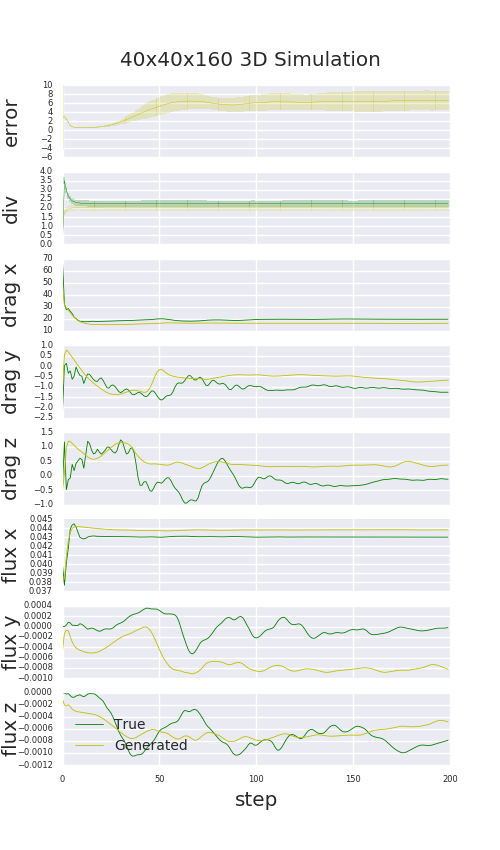}}
\subfigure{\includegraphics[scale=0.280]{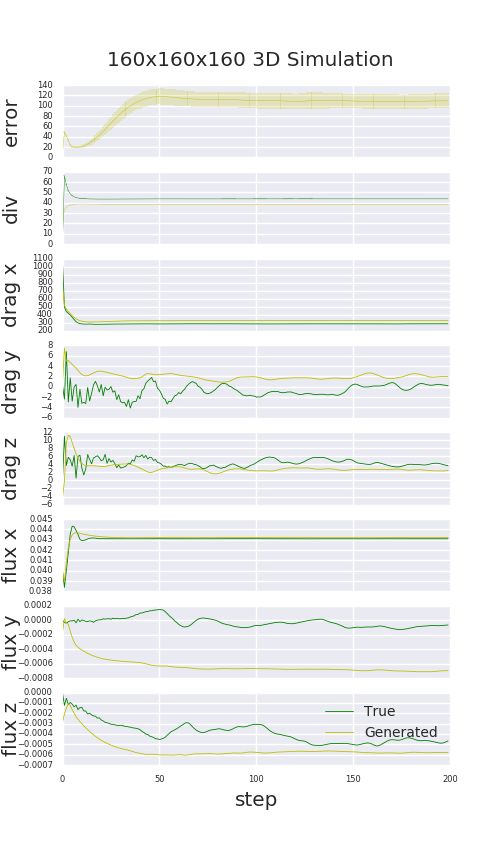}}
\caption{ Comparison plot of the flows generated by Lat-Net and the Lattice Boltzmann method. Each plot shows the average mean squared error of the true and generated generated along with the average divergence of the velocity vector field for both simulations. The standard deviation is also displayed. In addition, the calculated values for drag and flux are displayed for a single simulation run. For the 1024 by 1024 simulation, the flow produced by the Lattice Boltzmann solver tended to produce instabilities resulting in the chaotic divergence observed.}
\label{2d_error_plot}
\end{figure}

\begin{figure}[!t]
\centering
\subfigure{\includegraphics[scale=0.74]{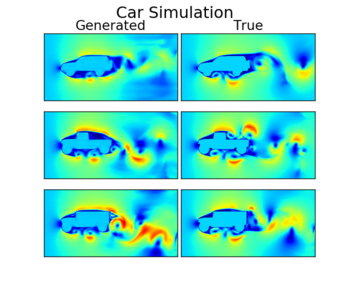}}
\subfigure{\includegraphics[scale=0.28]{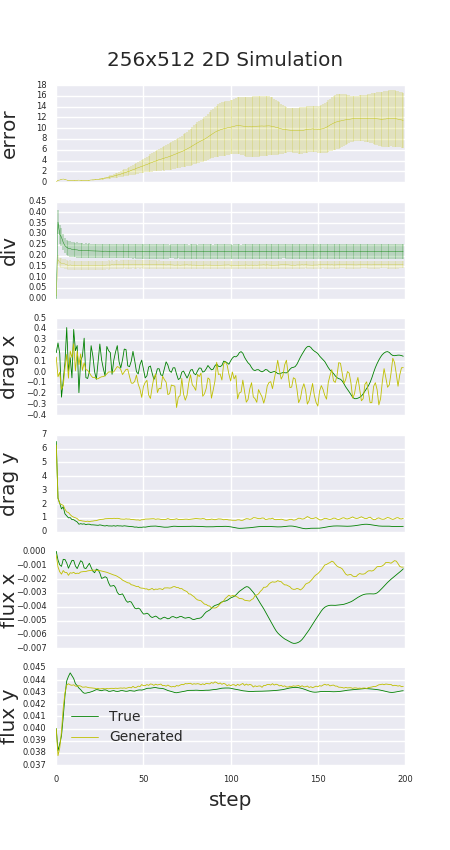}}
\caption{Comparison of generated flows for the vehicle cross section dataset. The images show the generated and true flow at step 100 for 3 different cars in the dataset. The plot shows the same values as described in \ref{2d_error_plot}.}
\label{car_dataset}
\end{figure}

A key component of our model is its ability to generate larger simulations then those trained on. To test its effectiveness in doing so, we compare the accuracy of generated 2D and 3D simulations to ground truth simulations.

Comparing the accuracy of our simulations require some consideration. The naive approach is to compare the MSE between the generated and true simulation at various time-steps. The problem with this approach is that fluid flow is a chaotic dynamical system and small perturbations in flow quickly compound leading to dramatic differences at latter times. For this reason, we compare a variety of metrics in evaluating the generated flows accuracy. Similar to \cite{tompson2016accelerating}, we compare the divergence of the generated and true velocity vector field to test our models stability. We also compare the computed values of drag and flux. Flow simulations are often run to calculate such values so comparing this is a strong indicator of our models real world applicability. The drag is calculated directly from the lattice state via the momentum transfer method \cite{guo2013lattice}. The flux value is the average of the flux in each non-boundary cell. These values can be used to calculate important quantities such as the drag coefficient and Reynolds number. Lastly, we visually inspect the produced flow to check for instabilities and blurring effects.

In figure \ref{2d_error_plot}, we see the predicted values for different grid sizes in 2D and 3D simulations. In the 2D simulations, Lat-Net is able to effectively transfer to larger domain sizes with very similar calculated values of drag and flux. The generated flow also maintains its stability even after hundreds of steps. In the 3D simulation we see that, while our model predicts realistic values for the 40x40x160 simulation, it tends to have a slight bias in the direction of flow that manifests itself in the 160x160x160 simulation.

When visually inspecting our produced flows (figure \ref{2d_image_plot}), we see a slight blurring effect but overall similar structure in the 2D flows. We attribute this blur effect to the dimensionality reduction and use of MSE. This can possibly be overcome with the use of generative adversarial network \cite{goodfellow2014generative} where the loss is derived from a discriminator network. Another solution may be to craft a loss function that takes advantage of the statistical properties of flow \cite{kim2008wavelet}. We leave these pursuits to future work.

There is a distinct difference in the generated and true flow for the 3D 160x160x160 simulation. While the generated flow appears accurate close to the objects, in regions far between objects the network tends to underestimate the flow velocity. We believe this is due to these types of regions not being present in the train set and is the probable cause for the biases seen in the drag and flux. As mentioned above, our 3D train set is limited due to memory constraints and so developing a diverse train set to overcome this proved difficult.

The boundaries used in the above evaluation are drawn from the same distribution as the train set. This motivates the question of how our model performs on drastically different geometries. To test this, we apply our model to predicting flow around vehicle cross sections. Surprisingly, even though are model is only trained on flows around simple shapes (ovals and rectangles) it can effectively generalize to this distinctly different domain. In figure \ref{car_dataset}, we see the predicted flows are quite similar but with the same blurring effect. Calculating the same values as above, we see the flow is stable and produces similar drag and flux.

\subsection{Computation and Memory Compression}

\begin{table}[]
\small
\caption{Computation Time of Network} \label{compute_times}
\centering
\begin{tabular}{|c|cccccc|}
\hline
Simulation    & Comp. Size       & Comp. Mapping       & Full State  & Plane      & Line       & Point \\ \hline
(1024, 1024, 9)  & (64, 64, 128)                 & 2.7 ms            & 36.2 ms   & NA         & 6.7 ms   & 6.6 ms \\
(160, 160, 160, 15) & (40, 40, 40, 64)                 & 23.1 ms           & 272.1 ms  & 38.2 ms  & 25.6 ms  & 24.1 ms  
\\ \hline
\end{tabular}
\label{computation_table}
\end{table}

In this section, we investigate the computational speed-up of our model. The standard performance metric for Lattice Boltzmann Codes is Million Lattice Updates per Second (MLUPs). This metric is calculated by the following equation,
\begin{equation}
  MLUP = \frac{n_x \times n_y \times n_z \times 10^{-6}}{Compute \ Time}
\end{equation}
 where $n_x$, $n_y$, and $n_z$ are the dimensions of the simulation. For 3 dimensional simulations like the ones seen in this paper, a speed of 1,200 MLUPS can be achieved with a Nvidia K20 GPU and single precision floats \cite{januszewski2014sailfish}. We use this as our benchmark value to compare against.

The computation time and memory usage of the encoder can be neglected because this is a one time cost for the simulation. In addition, if the simulation is started with uniformly initialized flow as seen in our experiments, the computation to compress the flow is extremely redundant and can easily be optimized.

As seen in table \ref{computation_table}, the computation time of the compression mapping is 23.1 ms for a 3D simulation of grid size 160 by 160 by 160. Because each step of the compression mapping is equivalent to 60 Lattice Boltzmann steps, this equates to 10,600 MLUPS and a roughly 9x speed increase (a similar speed-up is seen with the 2D simulation). This does not give a complete picture though. Once the compressed states have been generated, the flow must be extracted with the decoder. Unfortunately, this requires considerable amounts of computation and memory because it involves applying convolutions to the full state size. Fortunately, there are ways around this. In many applications of CFD, it is not necessary to to have the full state information of the flow at each time-step. For example, calculating the drag only requires integrating over the surface of the object. By using the convolutional nature of the decoder, we can extract specif pieces of the flow without needing to compute the full state. In table \ref{computation_table}, we show computation times for extracting flow information of a plane, line and single point. While these computations can still be somewhat expensive, they do not necessarily need to be performed at every time-step and require very little working memory. 

Unfortunately, there are some measurements that do require the full state information to compute such as the average flux seen in our tests. Our method is currently unable to handle these without requiring high run-times and large working memory. A possible solution is training a separate neural network that takes in the compressed state and predicts the desired measurement. This would negate the need to extract out the full state and keep memory usage low. We leave this and similar ideas for future work.

While Lat-Net compresses the simulation state size by more then an order of magnitude, the working memory requirements for the compression network must be considered. A typical GPU based Lattice Boltzmann solver requires around 1.5 times as much working GPU memory as the memory size of the lattice \cite{januszewski2014sailfish}. For example, the maximum sized D3Q15 lattice that can fit on a GPU with 8 Gigabytes is $446^3$. We have observed that in our implementation of Lat-Net, the maximum 3D compression network we can run with an 8 Gigabyte GPU corresponds to a lattice size of $672^3$. This represents a 3.4x efficiency gain in working memory usage. While this is certainly a nontrivial gain, we feel that further improvements can be realized with a more memory efficient implementation of the compression mapping.

\subsection{Electromagnetic Results}

\begin{figure}[!t]
\centering
\subfigure{\includegraphics[scale=0.28]{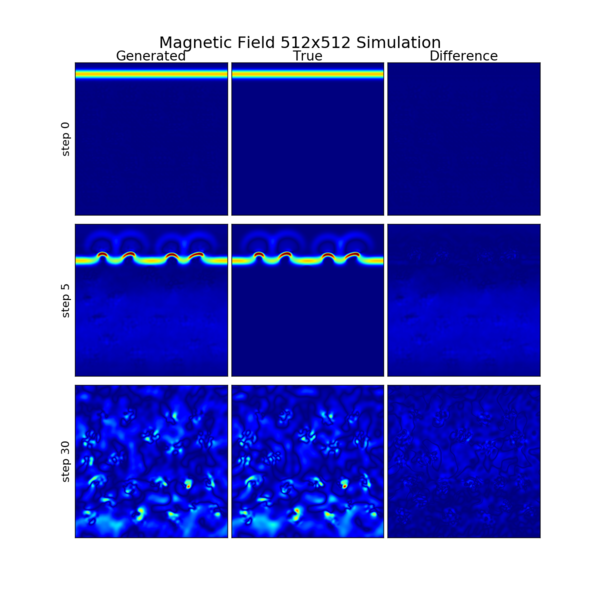}}
\subfigure{\includegraphics[scale=0.14]{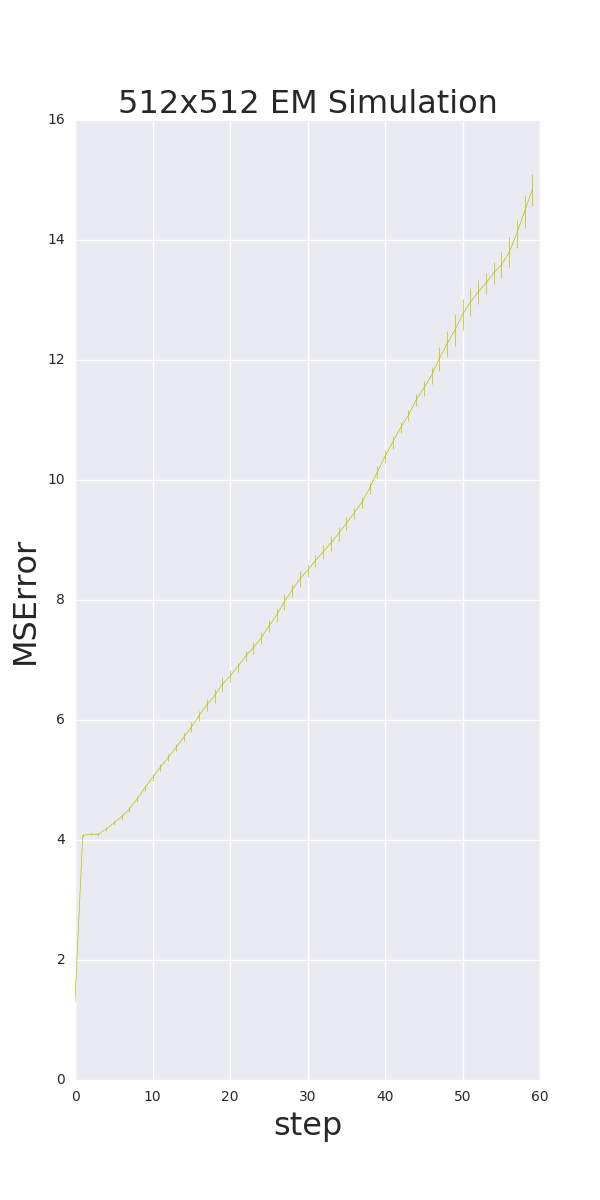}}
\caption{ Comparison of generated Electromagnetic fields. The images show the true and generated magnetic field a various time-steps. The reflection and refraction phenomena can clearly be seen in both. The plot shows the mean squared error of the true and generated simulation.}
\label{em_dataset}
\end{figure}

Finally, we illustrate the generality of our method by applying it to electromagnetic simulations. The same neural network architecture is used as in the 2D flow simulations with the only difference being the filter size on the compression is half of that in the flow network. The loss is kept identical however the lattice values are scaled up by a factor of 10 so they are on the same range as the flow lattice values. In figure \ref{em_dataset}, we see very similar waves formed with the same reflection and refraction.

\section{Conclusion}

Fluid Simulations are incredibly important for a variety of tasks however they are extremely computation and memory intensive. In this work, we have developed a unique method to overcome this using deep neural networks. We have demonstrated it is capable of accurately reconstructing a variety of simulations under different conditions with significantly less computation and memory. We have also shown that our method can be readily applied to other physics simulations such as electromagnetic simulations. While our method has proved successful on the problems in this paper, there is still significant room for improvement. A loss function that either takes into account the statistical nature of the flow or uses recent advances in GANs could produce shaper, more realistic flow. Training a network to extract desired measurements from the compressed state such as average flux would overcome the current memory limitation for such a task. We leave these and other improvements for future work.

\bibliography{references}

\begin{thebibliography}{10}

\bibitem{tompson2016accelerating}
J.~Tompson, K.~Schlachter, P.~Sprechmann, and K.~Perlin, ``Accelerating
  eulerian fluid simulation with convolutional networks,'' {\em arXiv preprint
  arXiv:1607.03597}, 2016.

\bibitem{guo2016convolutional}
X.~Guo, W.~Li, and F.~Iorio, ``Convolutional neural networks for steady flow
  approximation,'' in {\em Proceedings of the 22nd ACM SIGKDD International
  Conference on Knowledge Discovery and Data Mining}, pp.~481--490, ACM, 2016.

\bibitem{yang2016data}
C.~Yang, X.~Yang, and X.~Xiao, ``Data-driven projection method in fluid
  simulation,'' {\em Computer Animation and Virtual Worlds}, vol.~27, no.~3-4,
  pp.~415--424, 2016.

\bibitem{mcnamara1988use}
G.~R. McNamara and G.~Zanetti, ``Use of the boltzmann equation to simulate
  lattice-gas automata,'' {\em Physical review letters}, vol.~61, no.~20,
  p.~2332, 1988.

\bibitem{galindolattice}
S.~Galindo-Torres, A.~Scheuermann, and R.~Puscasu, ``A lattice boltzmann solver
  for maxwell equations in dielectric media,''

\bibitem{mendoza2010three}
M.~Mendoza and J.~Munoz, ``Three-dimensional lattice boltzmann model for
  electrodynamics,'' {\em Physical Review E}, vol.~82, no.~5, p.~056708, 2010.

\bibitem{kim2008wavelet}
T.~Kim, N.~Th{\"u}rey, D.~James, and M.~Gross, ``Wavelet turbulence for fluid
  simulation,'' in {\em ACM Transactions on Graphics (TOG)}, vol.~27, p.~50,
  ACM, 2008.

\bibitem{zhong2006lattice}
L.~Zhong, S.~Feng, P.~Dong, and S.~Gao, ``Lattice boltzmann schemes for the
  nonlinear schr{\"o}dinger equation,'' {\em Physical Review E}, vol.~74,
  no.~3, p.~036704, 2006.

\bibitem{shan1993lattice}
X.~Shan and H.~Chen, ``Lattice boltzmann model for simulating flows with
  multiple phases and components,'' {\em Physical Review E}, vol.~47, no.~3,
  p.~1815, 1993.

\bibitem{onodera2013large}
N.~Onodera, T.~Aoki, T.~Shimokawabe, and H.~Kobayashi, ``Large-scale les wind
  simulation using lattice boltzmann method for a 10 km$\times$ 10 km area in
  metropolitan tokyo,'' {\em TSUBAME e-Science Journal Global Scientific
  Information and Computing Center}, vol.~9, pp.~1--8, 2013.

\bibitem{xian2011multi}
W.~Xian and A.~Takayuki, ``Multi-gpu performance of incompressible flow
  computation by lattice boltzmann method on gpu cluster,'' {\em Parallel
  Computing}, vol.~37, no.~9, pp.~521--535, 2011.

\bibitem{goh2017deep}
G.~B. Goh, N.~O. Hodas, and A.~Vishnu, ``Deep learning for computational
  chemistry,'' {\em Journal of Computational Chemistry}, 2017.

\bibitem{mills2017deep}
K.~Mills, M.~Spanner, and I.~Tamblyn, ``Deep learning and the schr$\backslash$"
  odinger equation,'' {\em arXiv preprint arXiv:1702.01361}, 2017.

\bibitem{carleo2017solving}
G.~Carleo and M.~Troyer, ``Solving the quantum many-body problem with
  artificial neural networks,'' {\em Science}, vol.~355, no.~6325,
  pp.~602--606, 2017.

\bibitem{2017arXiv170502355P}
M.~{Paganini}, L.~{de Oliveira}, and B.~{Nachman}, ``{CaloGAN: Simulating 3D
  High Energy Particle Showers in Multi-Layer Electromagnetic Calorimeters with
  Generative Adversarial Networks},'' {\em ArXiv e-prints}, May 2017.

\bibitem{goodfellow2014generative}
I.~Goodfellow, J.~Pouget-Abadie, M.~Mirza, B.~Xu, D.~Warde-Farley, S.~Ozair,
  A.~Courville, and Y.~Bengio, ``Generative adversarial nets,'' in {\em
  Advances in neural information processing systems}, pp.~2672--2680, 2014.

\bibitem{rowley2004model}
C.~W. Rowley, T.~Colonius, and R.~M. Murray, ``Model reduction for compressible
  flows using pod and galerkin projection,'' {\em Physica D: Nonlinear
  Phenomena}, vol.~189, no.~1, pp.~115--129, 2004.

\bibitem{barone2009reduced}
M.~F. Barone, I.~Kalashnikova, M.~R. Brake, and D.~J. Segalman, ``Reduced order
  modeling of fluid/structure interaction,'' {\em Sandia National Laboratories
  Report, SAND No}, vol.~7189, pp.~44--72, 2009.

\bibitem{veroy2005certified}
K.~Veroy and A.~Patera, ``Certified real-time solution of the parametrized
  steady incompressible navier--stokes equations: rigorous reduced-basis a
  posteriori error bounds,'' {\em International Journal for Numerical Methods
  in Fluids}, vol.~47, no.~8-9, pp.~773--788, 2005.

\bibitem{rowley2005model}
C.~W. Rowley, ``Model reduction for fluids, using balanced proper orthogonal
  decomposition,'' {\em International Journal of Bifurcation and Chaos},
  vol.~15, no.~03, pp.~997--1013, 2005.

\bibitem{kingma2013auto}
D.~P. Kingma and M.~Welling, ``Auto-encoding variational bayes,'' {\em arXiv
  preprint arXiv:1312.6114}, 2013.

\bibitem{watter2015embed}
M.~Watter, J.~Springenberg, J.~Boedecker, and M.~Riedmiller, ``Embed to
  control: A locally linear latent dynamics model for control from raw
  images,'' in {\em Advances in Neural Information Processing Systems},
  pp.~2746--2754, 2015.

\bibitem{guo2013lattice}
Z.~Guo and C.~Shu, {\em Lattice Boltzmann method and its applications in
  engineering}, vol.~3.
\newblock World Scientific, 2013.

\bibitem{vondrick2016generating}
C.~Vondrick, H.~Pirsiavash, and A.~Torralba, ``Generating videos with scene
  dynamics,'' in {\em Advances In Neural Information Processing Systems},
  pp.~613--621, 2016.

\bibitem{he2016deep}
K.~He, X.~Zhang, S.~Ren, and J.~Sun, ``Deep residual learning for image
  recognition,'' in {\em Proceedings of the IEEE Conference on Computer Vision
  and Pattern Recognition}, pp.~770--778, 2016.

\bibitem{mathieu2015deep}
M.~Mathieu, C.~Couprie, and Y.~LeCun, ``Deep multi-scale video prediction
  beyond mean square error,'' {\em arXiv preprint arXiv:1511.05440}, 2015.

\bibitem{kingma2014adam}
D.~Kingma and J.~Ba, ``Adam: A method for stochastic optimization,'' {\em arXiv
  preprint arXiv:1412.6980}, 2014.

\bibitem{mechsys}
D.~Pedroso, R.~Durand, and S.~Galindo, ``Mechsys, multi-physics simulation
  library,'' 2015.

\bibitem{januszewski2014sailfish}
M.~Januszewski and M.~Kostur, ``Sailfish: A flexible multi-gpu implementation
  of the lattice boltzmann method,'' {\em Computer Physics Communications},
  vol.~185, no.~9, pp.~2350--2368, 2014.

\end{thebibliography}
\bibliographystyle{ieeetr}

\end{document}